\documentclass{article} 
\usepackage{iclr2025_conference,times}
\let\cite\citep

\usepackage{wrapfig}
\usepackage{lipsum} 
\usepackage{wrapfig}
\usepackage{xcolor}
\usepackage{colortbl}

\usepackage{amsmath,amsfonts,bm}
\usepackage{multicol}

\def\eqref#1{equation~\ref{#1}}

\def\1{\bm{1}}

\DeclareMathAlphabet{\mathsfit}{\encodingdefault}{\sfdefault}{m}{sl}
\SetMathAlphabet{\mathsfit}{bold}{\encodingdefault}{\sfdefault}{bx}{n}

\usepackage{microtype}
\usepackage{graphicx}
\usepackage{multirow}
\usepackage{booktabs}
\usepackage{colortbl}
\usepackage{subfloat}
\usepackage{floatrow}
\usepackage{float}
\usepackage{pifont}
\usepackage{wrapfig}
\usepackage{amssymb}

\newfloatcommand{capbtabbox}{table}[][\FBwidth]
\newfloatcommand{capbfigbox}{figure}[][\FBwidth]

\usepackage{algorithm}
\usepackage{algorithmicx}
\usepackage[noend]{algpseudocode}

\usepackage{pifont}
\usepackage{bbding}
\usepackage{caption}

\definecolor{uclablue}{rgb}{0.15, 0.45, 0.68}
\hypersetup{
    breaklinks,
    citecolor=uclablue,
    colorlinks=true,
}

\NewDocumentCommand{\xx}
{ mO{} }{\textcolor{blue}{\textsuperscript{\textit{todo}}\textsf{\textbf{\small[#1]}}}}
\usepackage{xspace}

\newcommand{\ours}{\textbf{\texttt{UI-Voyager}}\xspace}

\definecolor{greenbg}{RGB}{230, 255, 230}

\def\shownotes{1}  
\ifnum\shownotes=1
\newcommand{\authnote}[2]{[#1: #2]}
\else
\newcommand{\authnote}[2]{}
\fi

\usepackage{xcolor}
\usepackage{colortbl}

\title{\ours: A Self-Evolving GUI Agent Learning via Failed Experience}

\author{Zichuan Lin$^{*\dagger}$, Feiyu Liu$^*$, Yijun Yang$^*$, Jiafei Lyu$^*$, Yiming Gao$^*$, Yicheng Liu$^*$, Zhicong Lu \\ Yangbin Yu, Mingyu Yang, Junyou Li, Deheng Ye$^\ddagger$, Jie Jiang$^\ddagger$ \\
{\tt\small lzcthu12@gmail.com, ydyl1991@gmail.com, zeus@tencent.com} \\
\textbf{Tencent Hunyuan} \\
\textbf{Code and Models:} \href{https://github.com/ui-voyager/UI-Voyager}{github.com/ui-voyager/ui-voyager}
}

\begin{document}
\maketitle
\renewcommand*{\thefootnote}{\fnsymbol{footnote}}
\footnotetext{$^*$Equal contribution. $^\dagger$Project Lead. $^\ddagger$Corresponding Author.}
\begin{abstract}

Autonomous mobile GUI agents have attracted increasing attention along with the advancement of Multimodal Large Language Models (MLLMs). However, existing methods still suffer from inefficient learning from failed trajectories and ambiguous credit assignment under sparse rewards for long-horizon GUI tasks. To that end, we propose \ours, a novel two-stage self-evolving mobile GUI agent. In the first stage, we employ Rejection Fine-Tuning (RFT), which enables the continuous co-evolution of data and models in a fully autonomous loop. The second stage introduces Group Relative Self-Distillation (GRSD), which identifies critical fork points in group rollouts and constructs dense step-level supervision from successful trajectories to correct failed ones. Extensive experiments on AndroidWorld show that our 4B model achieves an 81.0\% Pass@1 success rate, outperforming numerous recent baselines and exceeding human-level performance. Ablation and case studies further verify the effectiveness of GRSD. Our method represents a significant leap toward efficient, self-evolving, and high-performance mobile GUI automation without expensive manual data annotation. 
\end{abstract}




\begin{figure}[h]
    \centering
    \includegraphics[width=0.85\linewidth]{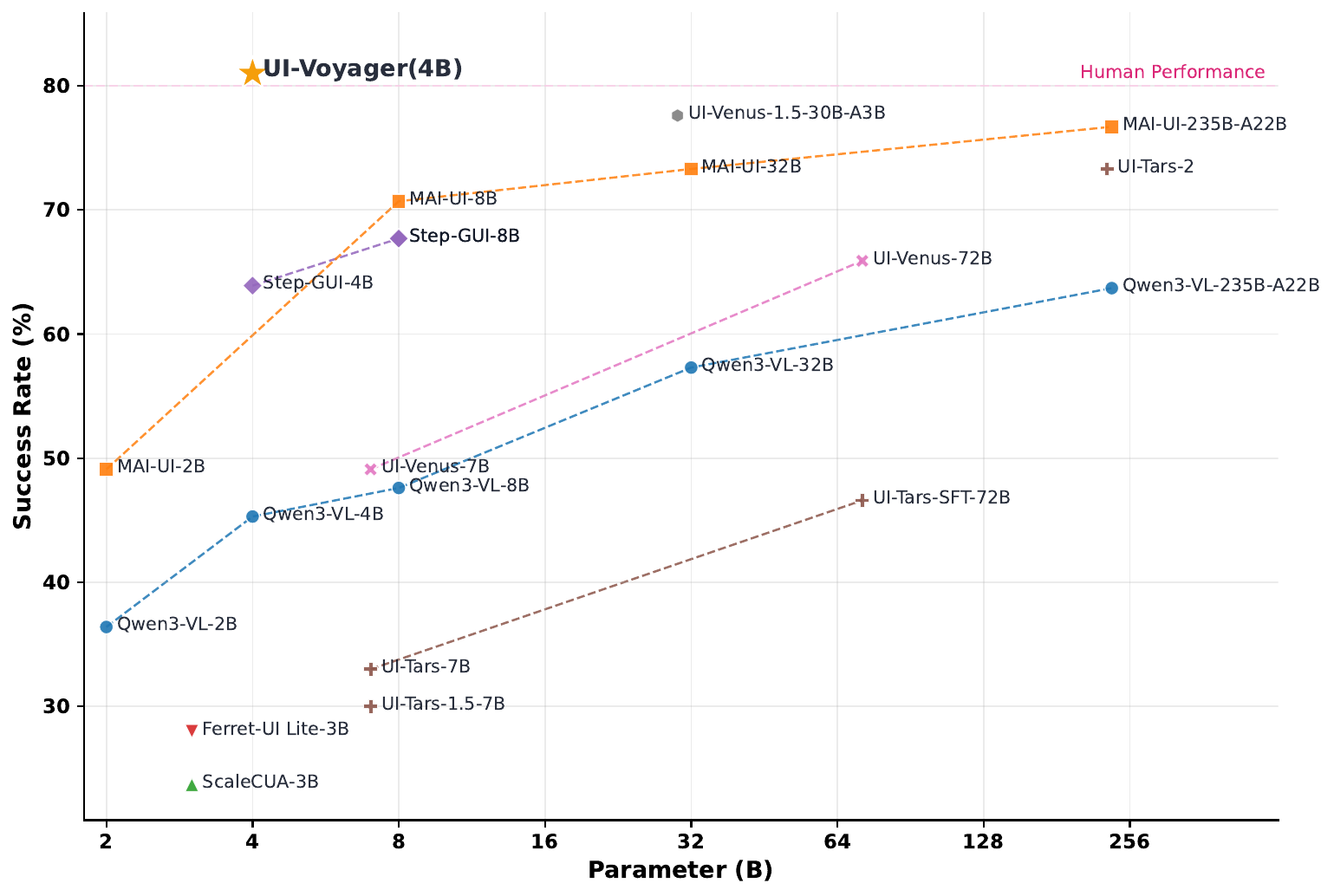}
    \caption{\textbf{Performance comparison of various GUI agents on AndroidWorld.} Our UI-Voyager (4B) achieves an 81.0\% Pass@1 success rate, outperforming larger models and exceeding reported human-level performance.}
    \label{fig:intro_fig}
\end{figure}

\section{Introduction}


The autonomous operation of intelligent digital systems, such as mobile phones, has been a long-standing pursuit and challenge. Some prior agents, like Siri (Apple) and Cortana (Microsoft), can only complete some predefined or simple operations. With the rapid development of Multimodal Large Language Models (MLLMs) \citep{bai2025qwen2,Bai2025Qwen3VLTR,Du2025KimiVLTR,Guo2025Seed15VLTR,Yang2025MMaDAML,Cao2025HunyuanImage3T,lin2025adaptvision} in recent
years, GUI agents \citep{deng2024multi,wang2024ant,chen2025guicourse,chen2025guiworld,lu2025guiodyssey} have emerged as a promising direction towards building generic, human-like intelligent agents capable of perceiving, understanding, planning, reasoning, and operating graphical user interfaces in a fully autonomous manner.


Among various GUI scenarios, mobile interfaces stand out as a representative and challenging domain \citep{rawles2025androidworld,chai2025a} due to their diverse screen layouts (the screen layout can be personalized), rich interaction styles (e.g., click, swipe, open various apps, input text), limited visual context, and dynamic state transitions. It is necessary and meaningful to study mobile GUI agents, considering the growing importance of mobile phones in people's daily lives. In fact, there have been numerous efforts to integrate strong MLLMs into mobile phones to build powerful mobile GUI agents \citep{Xu2025MobileRLOA,Ye2025MobileAgentv3FA,Shi2025MobileGUIRLAM,dai2025advancing,kang2026learning}, and there are already some practical applications, e.g., leveraging Doubao to operate the phone and using Qwen to order takeout. Despite remarkable progress in general GUI agents, mobile-oriented agents still suffer from the following issues: (i) \emph{inefficient learning upon failed trajectory.} During mobile interactions, failed trajectories constitute a large proportion of agent experience (especially on hard tasks), yet they are typically underutilized in conventional training pipelines, which limits data efficiency; (ii) \emph{ambiguous credit assignment} of existing Reinforcement Learning (RL) algorithms for the sparse reward case. The coarse-grained, trajectory-level rewards (success/failure) obtained from mobile GUI interactions make the agent incapable of identifying which specific step caused task failure, thus hindering stable policy optimization.

In light of the challenges above, we propose \ours in this work, a novel GUI agent trained via a two-stage self-evolving optimization pipeline. In the first stage, we employ the \textbf{Rejection Fine-Tuning (RFT)} strategy, which iteratively collects, filters, and refines GUI interaction trajectories without manual annotation, enabling automatic co-evolution of both training data and model capabilities. In the latter stage, we adopt the \textbf{Group Relative Self-Distillation (GRSD)} method to alleviate the severe credit assignment issue in long-horizon GUI tasks. GRSD identifies shared states (fork points) among group rollouts and extracts dense, step-level supervision from successful trajectories to supervise failed ones, which effectively replaces sparse trajectory-level rewards with precise self-distillation learning signals, reuses the failed trajectories, and mitigates the credit assignment issue.


To validate the effectiveness of the proposed \ours framework, we conduct experiments on the AndroidWorld \citep{rawles2025androidworld} benchmark, which features diverse tasks (116 tasks), easy-to-use evaluation protocol, and varying complexities across numerous real-world apps. Empirical results show that our 4B model achieves a Pass@1 success rate of \textbf{81.0\%}, surpassing all baseline methods and the reported human-level performance on AndroidWorld tasks. Further ablation studies and case studies confirm the critical contributions of the core components introduced in \ours. Specifically, we demonstrate how fork point detection works and the effectiveness of GRSD by comparing it against methods like GRPO. These results clearly show that \ours effectively mitigates the learning inefficiency issue and the credit assignment issue, moving a concrete step towards stronger and more powerful GUI agents. \looseness-1

\section{Related Work}
\label{sec:relatedwork}

\subsection{Interactive Environments}
\label{sec:interactiveenvironments}

For training GUI agents, many researchers resort to training the agent on large-scale static datasets that contain extensive interaction data collected from real app or web environments \citep{deng2023mindweb,rawles2023androidinthewild,cheng2024seeclick,deng2024multi,gao2024mobileviews,wang2024ant,chen2025guicourse,sun2025gui,chen2025guiworld,lu2025guiodyssey,chai2025amex}. This enables the agent to capture general GUI knowledge, such as action grounding, icon functionality, task decomposition, etc. However, the static nature of training with the datasets limits the agent's ability to handle unpredictable UI behaviors and learn from trial-and-error. In contrast, another line of research focuses on training and evaluating the GUI agent in interactive environments, which typically include the GUI interface of the computer desktop or mobile phone, and actions taken in the environment can alter the state \citep{nguyen2025gui}. There are numerous environments targeting web browsing \citep{shi2017world,liu2018learning,mialon2023gaia,zhang2026infiniteweb}, e.g., WebShop \citep{yao2022webshop}, WebArena \citep{zhou2024webarena}, VisualWebArena \citep{koh2024visualwebarena}, WorkArena \citep{drouin24workarena}, WebChoreArena \citep{miyai2025webchorearena}, etc. Some environments like OSWorld \citep{xie2024osworld,xie2025scaling}, WindowsAgentArena \citep{bonatti2024windows}, AgentStudio \citep{zheng2024agentstudio} are built for the purpose of general computer use. In the mobile domain, there are also many existing benchmarks, including Mobile-Env \citep{zhang2023mobile} and MobileWorld \citep{kong2025mobileworld}. The interactive environment can provide reward signals when the task is successfully completed \citep{abramson2022evaluating,ruan2024identifying,tian2025mmina}. In this work, we focus on the AndroidWorld \citep{rawles2025androidworld} environment, which involves 116 diverse and programmatic tasks with varying complexities and optimal interaction steps, making it a challenging benchmark for evaluating the performance of GUI agents.

\subsection{Interactive Agents}
\label{sec:interactiveagents}

Prior interactive agents are often emphasized in reinforcement learning (RL) where the agent interacts with the environment (e.g., game \citep{brockman2016openai,tassa2018deepmind,wei2022honor,wei2025gtr,wei2025gtr-turbo}, embodied setting \citep{puig2018virtualhome,savva2019habitat,yang2024embodied}) and optimizes the policy \citep{yang2019fully,lin2018episodic,lin2020model,lin2021juewu,lyu2022efficient,lyu2024off,lyu2024odrl}. Earlier trials in developing the UI-operating agent primarily use RL or behavior cloning to simulate interactions like mouse click \citep{shvo2021appbuddy,gur2021environment,Humphreys2022ADA}. With the advancement of foundation models, such as ChatGPT \citep{achiam2023gpt}, DeepSeek-R1 \citep{guo2025deepseek}, Qwen \citep{yang2025qwen3,wang2024qwen2,bai2025qwen2}, and Gemini \citep{team2023gemini}, existing large language models (LLMs) and large vision-language models (LVLMs) have led to significant breakthroughs in intent comprehension, multi-modal reasoning, and GUI understanding \citep{wei2022chain,you2024ferret,li2024ferret,hong2024cogvlm2,zhang2025critic,lin2025showui,liu2026textmonkey}. These models are now widely used in building strong GUI agents, either by leveraging these high-performing models for planning or by directly fine-tuning VLMs for downstream tasks \citep{xie2025gui,gu2025ui,luo2025gui,zhou2025gui,ye2025mobile,zeng2025uitron,huang2025spiritsight,wanyan2025look}. Interactive GUI agents are actively explored in many scenarios, including mobile phone \citep{yan2023gpt,bishop2024latent,zhang2024you,dai2025advancing,kang2026learning}, desktop OS \citep{Wu2024OSCopilotTG,zhang2025ufo,zhang2025ufo2,xie2024osworld,hu2025agents,zhang2026mirrorguard}, and desktop web \citep{zheng2024gpt,koh2024visualwebarena,cheng2024seeclick,song2025beyond,cai2025large,wei2025webagent}. Different from prior works, the focus of this work is to build a strong open-source interactive agent in AndroidWorld that can efficiently and successfully solve long-horizon, complex tasks. \looseness-1

To address the credit assignment problem~\citep{lu2026hisr} inherent in long-horizon GUI tasks, recent studies (e.g., EvoCUA \citep{xue2026evocua}) identify critical forking points and rely on external VLMs to synthesize correction traces for direct preference optimization~\citep{rafailov2023direct}. Targeting these crucial forking points, we propose a lightweight intra-group detection approach based on SSIM to locate divergent states without relying on any external models. Different from prior works, we introduce a Group Relative Self-Distillation (GRSD) mechanism that achieves robust policy improvement by directly distilling the correct actions of successful peer trajectories into the historical context of failed rollouts.

\section{Method}
\begin{figure}[htbp]
    \centering
    \includegraphics[width=\linewidth]{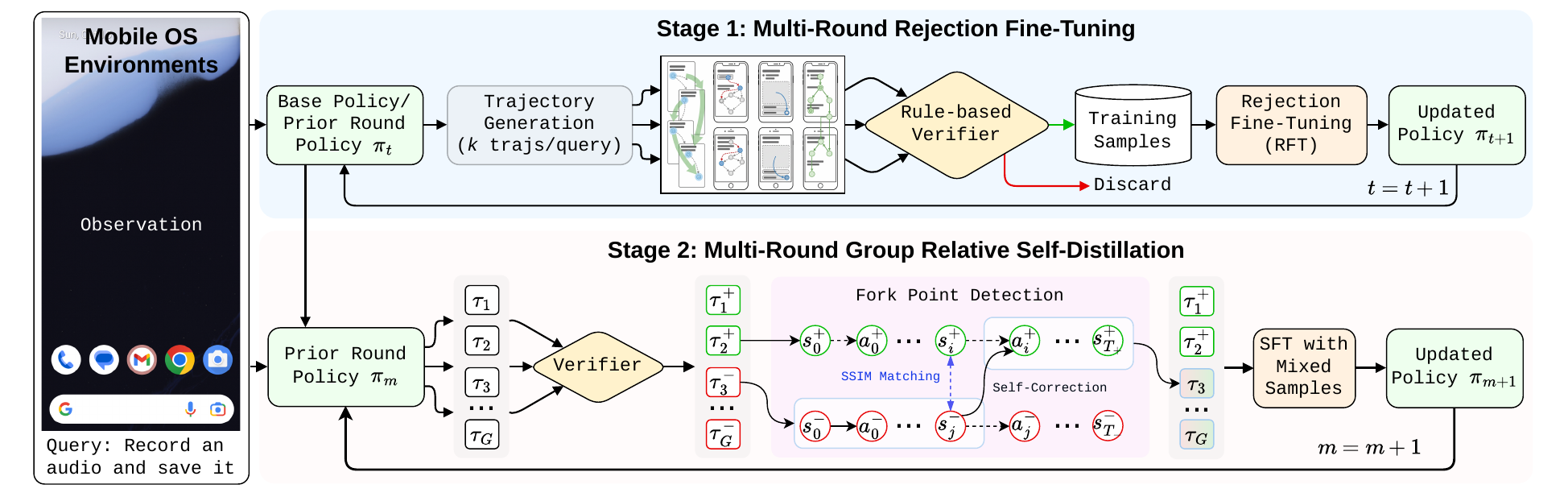}
    \caption{\textbf{The whole pipeline of training \ours for mobile GUI tasks.} It consists of two iterative stages: (1) Rejection Fine-Tuning (RFT), where a base policy generates multiple trajectories that are filtered by a rule-based verifier to collect high-quality samples for supervised fine-tuning; (2) Group Relative Self-Distillation (GRSD), which identifies ``fork points'' between successful and failed trajectory groups using SSIM matching and corrects erroneous actions to further refine the policy $\pi_m$ through mixed-data training.}
    \label{fig:framework}  
\end{figure}

\subsection{Overview}
We provide a comprehensive overview of \ours, including task formulation, the state and action spaces, and the agent architecture.

\paragraph{Task Formulation}
In the context of GUI tasks, the interaction is modeled as a Partially Observable Markov Decision Process (POMDP) defined by the tuple $(\mathcal{S}, \mathcal{O}, \mathcal{A}, \mathcal{T})$. Here, $\mathcal{S}$ represents the underlying states, while $\mathcal{O}$ constitutes the observation space, merging visual screenshots with linguistic instructions $\mathcal{I}$. The action space $\mathcal{A}$ encompasses common mobile UI interactions such as clicking, swiping, and typing, as listed in Table~\ref{tab:action_space}. The state transitions are given by $\mathcal{T}: \mathcal{S} \times \mathcal{A} \rightarrow \mathcal{S}$. At each step $t$, the agent determines its next move $a_t = \pi(\mathcal{I}, o_t, \mathcal{H}_t)$, where $\mathcal{I}$ is the task instruction, $o_t$ is the current observation, and $\mathcal{H}_t = (a_{t-h},o_{t-h}, ..., a_{t-1}, o_{t-1})$ denotes the history context of previous actions and observations with window size $h$. Task completion is determined by a durable, rule-based verifier, which assigns a shaped scalar reward $\mathcal{R}: \mathcal{S} \times \mathcal{A} \rightarrow \mathbb{R}$ by checking application states using the Android Debug Bridge (\texttt{adb command}).

\begin{table}[htb]
\caption{\textbf{Predefined action space of AndroidWorld~\cite{rawles2025androidworld}.}}
\label{tab:action_space}
\begin{center}
\begin{small}
\begin{tabular}{ll}
\toprule
\textbf{Code Actions} & \textbf{Descriptions}  \\
\midrule
\texttt{click(x,y)}            & Clicking at coordinates \texttt{(x,y)} \\
\texttt{long\_press(x,y)}      & Long-pressing at coordinates \texttt{(x,y)} \\
\texttt{swipe(x,y,x',y')}      & Swiping from \texttt{(x,y)} to \texttt{(x',y')} \\
\texttt{open\_app(app\_name)}  & Opening an app by name \\
\texttt{input\_text(text)}     & Typing input text \\
\texttt{keyboard\_enter()}     & Pressing the Enter key on the keyboard \\
\texttt{navigate\_back()}      & Pressing the system Back button \\
\texttt{navigate\_home()}      & Pressing the system Home button \\
\texttt{wait()}                & Waiting / no-op action (also used for unsupported actions) \\
\texttt{status(goal\_status)}  & Terminating the episode with status, e.g., \texttt{success} \\
\texttt{answer(text)}          & Returning the final answer text \\
\bottomrule
\end{tabular}
\end{small}
\end{center}
\end{table}

\paragraph{Agent Architecture} 

As shown in Fig. \ref{fig:framework}, \ours is trained via a two-stage
self-evolving optimization pipeline:
(1) Rejection Fine-Tuning (RFT), which employs a multi-round rejection sampling mechanism where trajectories generated by the prior policy are filtered by a rule-based verifier to collect high-quality training samples for iterative model updates; and 
(2) Group Relative Self-Distillation (GRSD), which identifies discrepancies between correct and incorrect trajectory groups through Fork Point Detection, enabling the model to learn from self-corrected transitions and achieve robust policy refinement.

\subsection{Rejection Fine-Tuning}
Similar to recent work~\cite{Yan2025StepGUITR,Zhou2025MAIUITR}, we employ a closed-loop self-evolving training pipeline to facilitate the mutual enhancement of training data and model capabilities, thereby improving GUI agent performance. This pipeline consists of two main modules: Trajectory Generation and Rejection Sampling.

\paragraph{Trajectory Generation.}
To provide diverse and novel task trajectories for both SFT and the subsequent GRSD stages, we design a seed task generator that synthesizes novel tasks by perturbing key parameters—such as temporal constraints, quantities, and file entities—from original task templates. Given the labor-intensive nature of human annotation, which is difficult to scale, we rely on GUI agents to automate trajectory synthesis. By combining automated execution in GUI environments with the task generator, we establish a high-throughput pipeline for generating diverse trajectories. This closed-loop paradigm fosters a co-evolutionary cycle in which model refinements and high-quality data synthesis reinforce each other.

\paragraph{Rejection Sampling.}
After generating diverse raw trajectories, we apply a rejection sampling mechanism to curate a high-fidelity SFT dataset. Only ``successful'' trajectories—those that either reach the predefined goal or pass a task-completion verifier—are retained. This rigorous filtering process ensures the structural integrity of the trajectories and the correctness of individual action steps, resulting in a refined, high-quality SFT corpus.

\paragraph{Iterative Training.}
In the initial iteration, we deploy various scales of the Qwen3-VL series as GUI agents for trajectory generation, using Qwen3-VL-4B-Instruct as the base model for SFT. In subsequent iterations, the model from the previous iteration serves as the agent to generate new trajectories. These trajectories are filtered through rejection sampling, and the resulting high-quality samples are used to fine-tune the model for the next round. Notably, each iteration uses new tasks generated by the seed task generator to maintain novelty and prevent overfitting.

\paragraph{Empirical Results.}
This self-evolving approach creates a synergy between data quality and model capability. Experimental results show that after three iterations, the Pass@1 score improves significantly from 37\% to 73\%, with consistent gains observed across all Pass@K metrics.

\subsection{Group Relative Self-Distillation}


During agentic RL (multi-turn) training, a natural choice is to adopt Group Relative Policy Optimization (GRPO)~\cite{shao2024deepseekmath} or Proximal Policy Optimization (PPO)~\cite{schulman2017proximal}. GRPO samples a group of responses $\{ o_i \}^G_{i=1}$ for each task $q$ and optimizes the policy via maximizing the objective below:
\begin{equation}
    \mathcal{J}_{GRPO} = \mathbb{E}_{q, o_i} \left[
        \frac{1}{\sum^G_{i=1} |o_i|} \sum_{i=1}^G \sum_{t=1}^{|o_i|} \textup{min}\Big(
            r_{i,t}(\theta) \hat{A}_{i,t}, \textup{clip} \Big( r_{i,t}(\theta), 1-\epsilon_{\textup{low}}, 1+\epsilon_{\textup{high}} \Big) \hat{A}_{i,t}
        \Big)
    \right],
\end{equation}
where $r_{i,t}(\theta) = \frac{ \pi_{\theta} (o_{i,t} \mid q, o_{i,<t}) }
{ \pi_{\theta_{old}} (o_{i,t} \mid q, o_{i,<t}) }$ is the token-level importance sampling ratio, and $\hat{A}_{i,t} = \frac{ R^{(i)} - \textup{mean}(\{R^{(i)}\}^G_{i=1}) }{ \textup{std}(\{R^{(i)}\}^G_{i=1}) }$ is the normalized advantage. In contrast to GRPO, which relies on group-based statistics to estimate the advantage, PPO does not require multiple rollouts per task. Instead, it utilizes a value network to estimate the value function, typically employing Generalized Advantage Estimation (GAE)~\cite{schulman2015GAE} to achieve a more accurate and variance-reduced estimate of the advantage function. The PPO objective is defined as:
\begin{equation}
\mathcal{J}_{PPO} = \mathbb{E}_{q, o} \left[
\frac{1}{|o|} \sum_{t=1}^{|o|} \textup{min}\Big(
r_{t}(\theta) \hat{A}_{t}^{GAE}, \textup{clip} \Big( r_{t}(\theta), 1-\epsilon, 1+\epsilon \Big) \hat{A}_{t}^{GAE}
\Big)
\right],
\end{equation}
where $r_{t}(\theta) = \frac{ \pi_{\theta} (o_{t} \mid q, o_{<t}) } { \pi_{\theta_{old}} (o_{t} \mid q, o_{<t}) }$ is the importance sampling ratio and $\hat{A}_{t}^{GAE}$ is the advantage estimated by GAE.

However, applying GRPO/PPO to multi-turn and long-horizon GUI agent training presents a fundamental challenge -- \textbf{credit assignment}. Since the reward $R_{\text{base}}$ is only assigned at the trajectory level: $1$ for success and $0$ for failure, and the advantage $\hat{A}_{i,t}$ is identical for every token within the same trajectory. The agent receives no signal about \emph{which step} caused the failure or \emph{what action} should have been taken instead. In tasks with up to 30 interaction steps, this trajectory-level reward makes learning extremely inefficient: a single wrong action at step 5 may cause a 30-step trajectory to receive zero reward, yet the other 29 correct actions also receive zero credit.


\textbf{Key insight.} When performing group rollouts for the same task, the $G$ trajectories often visit \emph{identical screen states} at certain steps but diverge due to different actions. These \textbf{fork points}---where the agent sees the same observation but makes a different decision---represent critical moments for step-level corrective supervision. Crucially, the successful trajectories within the same group can serve as \emph{teachers} for the failed ones: by identifying where they share the same state and how they diverge, we can extract precise, token-level supervision without any external annotation, as illustrated by Figure~\ref{fig:forkpointdetection}.

We formalize this idea as \textbf{Group Relative Self-Distillation (GRSD)}: within each group of $G$ rollouts, the shortest successful trajectory is selected as a ``teacher'', and its correct actions at fork points are distilled into the failed ``student'' trajectories via supervised fine-tuning. This transforms sparse trajectory-level feedback into dense step-level supervision, enabling targeted self-correction. GRSD differs from recent on-policy distillation (OPD) variants \citep{Lu2025OPD,Zhang2026FastAE,Zhao2026SelfDistilledRO,Xiong2026OVDOV} in that it enjoys a more concise, practical learning paradigm that does not depend on any explicit teacher policy and skillfully distills knowledge from self-generated successful trajectories through SFT.

\subsubsection{Fork Point Detection}
\label{sec:fpd}

\begin{figure}
    \centering
    \includegraphics[width=0.9\linewidth]{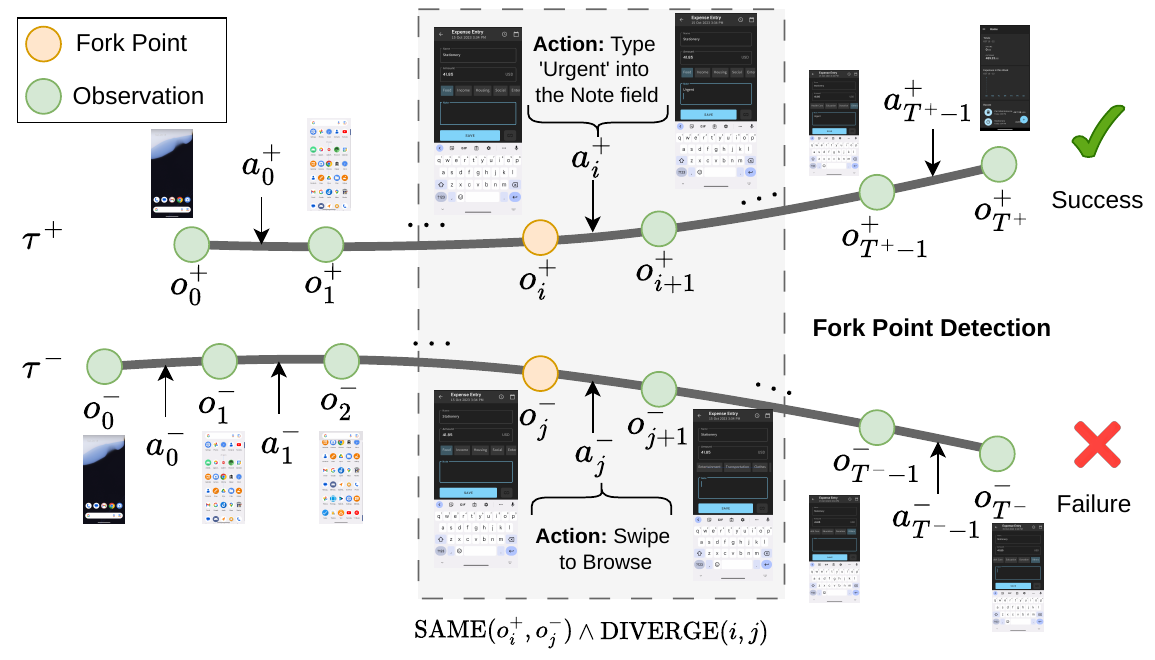}
    \caption{\textbf{Illustration of the fork point detection strategy.} Given a successful trajectory \(\tau^+\) and a failed trajectory \(\tau^-\) for the same task, the fork point detection mechanism identifies steps in the failed trajectory where the screen state matches that of a successful step (\(\text{SAME}(o_i^+, o_j^-)\)) but the subsequent action leads to divergence (\(\text{DIVERGE}(i, j)\)), indicating that the action taken in the failed trajectory deviates from the successful one. See Sec.~\ref{sec:fpd} for details.}
    \label{fig:forkpointdetection}
\end{figure}

We now describe how to extract step-level supervision from paired trajectories. Given a successful trajectory $\tau^+ = \{(o_0^+, a_0^+), \ldots, (o_{T^+}^+, a_{T^+}^+)\}$ and a failed trajectory $\tau^- = \{(o_0^-, a_0^-), \ldots, (o_{T^-}^-, a_{T^-}^-)\}$ for the same task, where $o_t$ is the screen observation (screenshot) and $a_t$ is the action taken at step $t$, our goal is to find \emph{fork points}: steps in the failed trajectory where the agent observed the same screen state as some step in the successful trajectory, yet chose a different—and ultimately wrong—action.

\paragraph{Cross-Trajectory State Matching.}
We define an observation equivalence function to determine whether two screenshots depict the same screen state. While a pretrained vision encoder could, in principle, be used to compute cosine similarity between visual embeddings, we opt for a more practical approach: Structural Similarity Index (SSIM)~\cite{brunet2011mathematical}. To accelerate computation, each screenshot is first cropped to remove a fixed-height status bar, resized to a low-resolution thumbnail, and converted to grayscale. A mean-hash pre-filter quickly discards obviously dissimilar pairs (hash similarity below 0.80) before the more expensive SSIM computation:
\begin{equation}
    \textsc{Same}(o_a, o_b) = \mathbb{1}\!\left[\text{SSIM}\!\left(\phi(o_a),\, \phi(o_b)\right) \geq \theta\right],
\end{equation}
where $\phi(\cdot)$ denotes the crop-resize-grayscale preprocessing pipeline and $\theta$ is the similarity threshold.


\paragraph{Transition Alignment}
Before matching a teacher step for failed step $j$, we perform a transition-alignment check: if there exists a successful step $i$ such that $\textsc{Same}(o_i^+, o_j^-)$ and $\textsc{Same}(o_{i+1}^+, o_{j+1}^-)$, we treat the trajectory prefixes as aligned. In this case, we skip failed step $j$ and advance the minimum successful index to $i_{\min} \leftarrow i+1$ for all subsequent failed steps $j'>j$.

\paragraph{Teacher Step Selection.}
For each remaining failed step $j$, we search over successful steps $i \ge i_{\min}$ to find the best \emph{teacher step}, subject to two conditions: (1) observation equivalence (i.e., $\textsc{Same}(o_i^+, o_j^-)$); 
and (2) transition divergence:
\begin{equation}
    \textsc{Diverge}(i, j) = \begin{cases}
        \textbf{true}  & \text{if } i = T^+ \text{ or } j = T^- \\
        \textbf{true}  & \text{if } \text{SSIM}(\phi(o_{i+1}^+),\, \phi(o_{j+1}^-)) < \theta \\
        \textbf{false} & \text{otherwise}
    \end{cases}
\end{equation}
If both trajectories have a subsequent step and the resulting observations are nearly identical, the two actions are considered to have the same effect and the pair is discarded as uninformative.

Among all qualifying teacher step candidates $\mathcal{C}(j)$, we select the one with the highest SSIM score, breaking ties by preferring the smallest successful-step index:
\begin{equation}
    i^*(j) = \arg\max_{i \in \mathcal{C}(j)} \left(\text{SSIM}(\phi(o_i^+),\, \phi(o_j^-)),\; -i\right).
\end{equation}

Crucially, we enforce a \emph{monotonicity constraint}: once a failed step $j$ is matched to a successful step $i^*(j)$, any subsequent failed step $j' > j$ can only match successful steps $i \ge i^*(j)$. This preserves the temporal ordering between the two trajectories, preventing pathological alignments where later failed steps map to earlier successful steps. Each failed step matches at most one successful step, but the same successful step may serve as the teacher for multiple failed steps. 

Figure \ref{fig:forkpointdetection} presents the general illustration of the fork point detection. Note that the fork point detection mechanism can also be extended to the language-only scenarios (e.g., observation $o$ is the text. We can discard $\phi(\cdot)$ and directly compute $\textsc{Sim}(o_i^+, o_j^-)$, where $\textsc{Sim}(\,\cdot\,,\,\cdot\,)$ is the similarity measure). Algorithm~\ref{alg:fork-point} summarized the fork point detection mechanism.

\begin{algorithm}[t]
\caption{Fork Point Detection}
\label{alg:fork-point}
\begin{algorithmic}[1]
\Require Successful trajectory $\tau^+$, failed trajectory $\tau^-$, threshold $\theta$
\Ensure Fork point set $\mathcal{M}$
\State $\mathcal{M} \leftarrow \emptyset$, $i_{\min} \leftarrow 0$

\For{$j = 0$ to $T^-$}
    \If{$\exists i \ge i_{\min}$ s.t. $\textsc{Same}(o_i^+, o_j^-)$ and $\textsc{Same}(o_{i+1}^+, o_{j+1}^-)$}
        \State $i_{\min} \leftarrow i+1$ \Comment{Transition Alignment}
        \State \textbf{continue}
    \EndIf

    \State $\mathcal{C}(j) \leftarrow \{\, i \ge i_{\min} \mid \textsc{Same}(o_i^+, o_j^-) \land \textsc{Diverge}(i,j)\,\}$
    \If{$\mathcal{C}(j)=\emptyset$} \State \textbf{continue} \EndIf

    \State $i^*(j) \leftarrow \arg\max_{i \in \mathcal{C}(j)} \big(\text{SSIM}(\phi(o_i^+),\phi(o_j^-)),-i\big)$ \Comment{Teacher Step Selection}
    \State $\mathcal{M} \leftarrow \mathcal{M} \cup \{(j,i^*(j))\}$
    \State $i_{\min} \leftarrow i^*(j)$
\EndFor

\State \Return $\mathcal{M}$
\end{algorithmic}
\end{algorithm}

\subsubsection{Step-Level Self Distillation}

For each identified fork point $(j, i^*(j))$, we construct a training sample by retaining the failed trajectory's prompt (including its contextual history at step $j$) and replacing the response with the successful trajectory's response at step $i^*(j)$:
\begin{equation}
    \mathbf{x}_j^{\text{train}} = \left[\underbrace{\text{prompt}_j^{-}}_{\text{failed-context prompt}} \;\middle|\; \underbrace{\text{response}_{i^*(j)}^{+}}_{\text{correct action}}\right].
\end{equation}

The training objective is the standard autoregressive next-token prediction loss computed \emph{only over the response tokens}:
\begin{equation}
    \mathcal{L}_{\text{GRSD}} = 
    -\frac{1}{|\mathcal{D}|} \sum_{\mathbf{x} \in \mathcal{D}} 
    \frac{1}{T_{\mathbf{x}}} \sum_{t=1}^{T_{\mathbf{x}}} 
    \log \pi_\theta\!\left(y_t \mid s_1, \ldots, s_{P_{\mathbf{x}}},\, y_{<t} \right),
\end{equation}
where $\mathcal{D}$ is the set of constructed samples, $s_{1:P_{\mathbf{x}}}$ are prompt tokens, $y_{1:T_{\mathbf{x}}}$ are response tokens, and $P_{\mathbf{x}}$ and $T_{\mathbf{x}}$ are prompt and response lengths, respectively.


In our experiments, we use GRSD as the sole training objective, replacing GRPO and PPO. This reflects the insight that for complex multi-step GUI tasks, precise step-level self-distillation from successful peers is more effective than trajectory-level advantage estimation with sparse rewards.

\section{Experiment}

\definecolor{tablepurple}{HTML}{E5E5FF}
\definecolor{headergray}{HTML}{F2F2F2}

\begin{table}[t] 
\centering
\caption{Performance comparison on \textbf{AndroidWorld} Benchmark. Best results are in \textbf{bold}, and second-best results are \underline{underlined}. 
\ours achieves an 81.0\% success rate, surpassing all baseline methods and the reported human-level performance of 80.0\%. Notably, our model achieves superior results with only 4B parameters, demonstrating strong efficiency compared to much larger models. To ensure reproducibility, we report the average success rate over 64 random seeds, whereas baseline results are taken from prior papers.}
\label{tab:main_result}
\small
\begin{tabular}{lcc}
\toprule
\textsc{\textbf{Model}} & \textsc{\textbf{\#Params}} & \textsc{\textbf{Success Rate}} \\ 
\midrule
\rowcolor{headergray} \textit{Baselines} & & \\
Qwen3-VL-2B \citep{Bai2025Qwen3VLTR} & 2B & 36.4 \\
MAI-UI-2B \citep{zhou2025mai} & 2B & 49.1 \\
ScaleCUA-3B \citep{liu2025scalecua} & 3B & 23.7 \\
Ferret-UI Lite-3B \citep{yang2025ferret} & 3B & 28.0 \\
Qwen3-VL-4B \citep{Bai2025Qwen3VLTR} & 4B & 45.3 \\
Step-GUI-4B \citep{Yan2025StepGUITR} & 4B & 63.9 \\
UI-Tars-7B \citep{qin2025ui} & 7B & 33.0 \\
UI-Tars-1.5-7B \citep{uitars15seed} & 7B & 30.0 \\
UI-Venus-7B \citep{gu2025ui} & 7B & 49.1 \\
GUI-Owl-7B \citep{ye2025mobile} & 7B & 66.4 \\
Step-GUI-8B \citep{yan2025step} & 8B & 67.7 \\
Qwen3-VL-8B \citep{Bai2025Qwen3VLTR} & 8B & 47.6 \\
MAI-UI-8B \citep{zhou2025mai} & 8B & 70.7 \\
Step-GUI-8B \citep{Yan2025StepGUITR} & 8B & 67.7 \\
GUI-Owl-1.5-8B-Thinking \citep{xu2026mobile} & 8B & 71.6 \\
UI-Venus-1.5-30B-A3B \citep{gao2026ui} & 30B & \underline{77.6} \\
Qwen3-VL-32B \citep{Bai2025Qwen3VLTR} & 32B & 57.3 \\
MAI-UI-32B \citep{zhou2025mai} & 32B & 73.3 \\
UI-Tars-SFT-72B \citep{qin2025ui} & 72B & 46.6 \\
UI-Venus-72B \citep{gu2025ui} & 72B & 65.9 \\
Seed1.5-VL \citep{Guo2025Seed15VLTR} & - & 62.1 \\
UI-Tars-2 \citep{wang2025ui} & 230B & 73.3 \\ 
Qwen3-VL-235B-A22B \citep{Bai2025Qwen3VLTR} & 235B & 63.7 \\
UI-Tars-1.5 \citep{uitars15seed} & - & 64.2 \\
Gemini-2.5-Pro \citep{gemini25pro} & - & 69.7 \\
Seed1.8 \citep{seed18} & - & 70.7 \\
MAI-UI-235B-A22B \citep{zhou2025mai} & 235B & 76.7 \\ 
\midrule
Human \citep{rawles2025androidworld} & - & 80.0 \\ 
\midrule
\rowcolor{headergray} \textit{Ours} & & \\
\ours & 4B & \textbf{81.0} \\ 
\bottomrule
\end{tabular}
\end{table}

\subsection{Experimental Setup}

\paragraph{Implementation details} \ours uses Qwen3-VL-4B-Instruct~\cite{Bai2025Qwen3VLTR} as the backbone. We evaluate on a popularly used Mobile GUI benchmark: AndroidWorld~\cite{rawles2025androidworld}, which comprises 116 diverse tasks across real-world mobile applications with varying complexity levels. 
AndroidWorld provides randomized initialization parameters, which enables the generation of a large number of training tasks with verifiable rewards by replacing predefined substitutable components in the tasks and by varying the initial device states. 
During training, we follow MobileRL~\cite{Xu2025MobileRLOA} and employ training sets from AndroidWorld, consisting over 7000 tasks. 

\paragraph{Baselines} Our baselines include both closed- and open-source agents and models, including Qwen-VL series~\citep{Bai2025Qwen3VLTR}, UI-Tars~\citep{qin2025ui,uitars15seed}, MAI-UI~\citep{Zhou2025MAIUITR}, Step-GUI~\citep{Yan2025StepGUITR}, MobileAgent~\citep{xu2026mobile,ye2025mobile}, Seed-VL series~\citep{Guo2025Seed15VLTR,seed18}, Gemini~\citep{gemini25pro}, UI-Venus~\citep{gu2025ui,gao2026ui}, ScaleCUA~\citep{liu2025scalecua}. 

\subsection{Main Results}

We report the pass@1 success rate and compare with a wide range of baseline models, including general-purpose VLMs (e.g., Qwen3-VL series), specialized GUI agents (e.g., UI-Tars, GUI-Owl, Step-GUI, MAI-UI, UI-Venus), and large-scale proprietary models (e.g., Gemini-2.5-Pro, Seed1.8).

As shown in Table~\ref{tab:main_result}, \ours (4B) achieves 81.0\% success rate, outperforming all baseline methods and surpassing the reported human-level performance of 80.0\% on AndroidWorld. Notably, our model achieves this with only \textbf{4B parameters}, demonstrating superior efficiency compared to much larger models such as MAI-UI-235B-A22B (76.7\%), UI-Tars-2 (73.3\%), and Qwen3-VL-235B-A22B (63.7\%). Even among models of comparable size, our approach significantly exceeds Step-GUI-4B (63.9\%) and Qwen3-VL-4B (45.3\%), highlighting the effectiveness of our training framework. To ensure reproducibility, we report the average success rate of \ours over 64 independent runs with randomized task parameters, whereas baseline results are taken from prior papers.

These results demonstrate that \ours, through its fork point detection and self-distillation mechanisms, effectively addresses the credit assignment challenge in long-horizon GUI agent learning, enabling a compact 4B model to achieve superior performance on AndroidWorld.

\subsection{Analysis}

In this part, we provide some case studies on the algorithmic components in \ours, including fork point detection and self-corrective samples. These analysis can provide more insights and help better understand the effectiveness of our propsoed \ours framework.

\begin{figure}[t]
    \centering
    \includegraphics[width=0.49\linewidth]{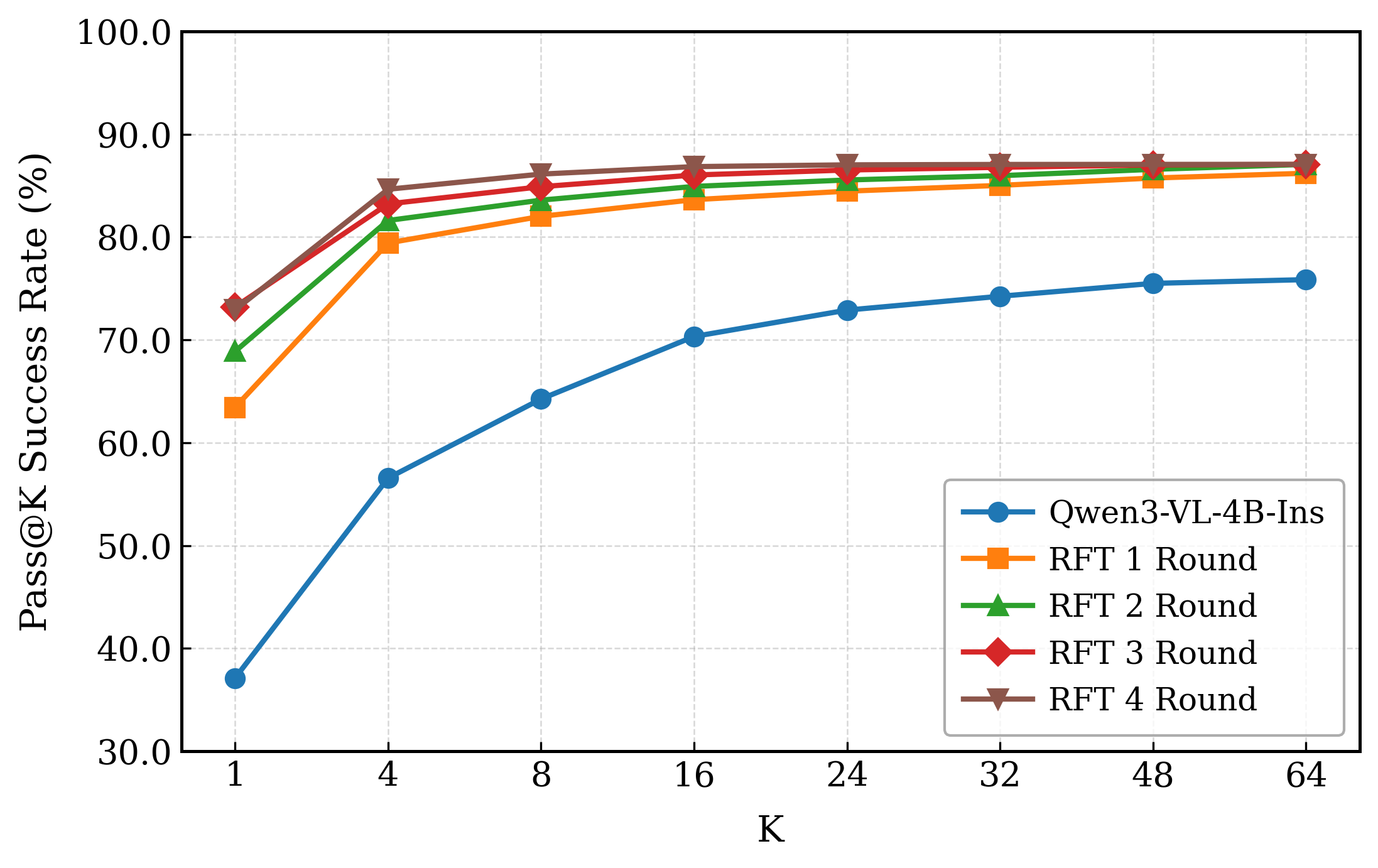}
    \includegraphics[width=0.49\linewidth]{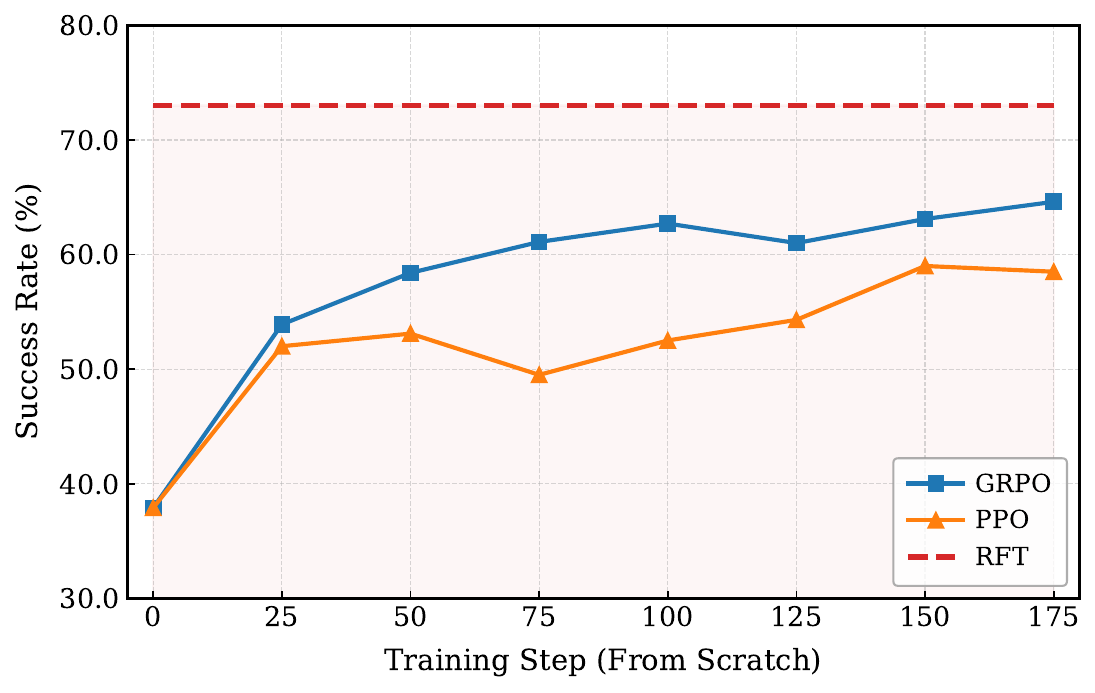}
    \caption{\textbf{RFT significantly boosts agent performance.} Left: Pass@K performance across four iterative rounds of RFT. The results show consistent improvement in both Pass@1 and Pass@k as the self-evolution progresses. We select the checkpoint from the third RFT round (Pass@1=73.2\%) for subsequent training. Right: Training curves of GRPO and PPO initialized from Qwen3-VL-4B-Instruct. The results show that directly deploying RL algorithms from Qwen3-VL-4B-Instruct model yields marginal gains and exhibits high sample inefficiency.}
    \label{fig:self_evolving}
\end{figure}

\begin{figure}[t]
    \centering
    \includegraphics[width=0.97\linewidth]{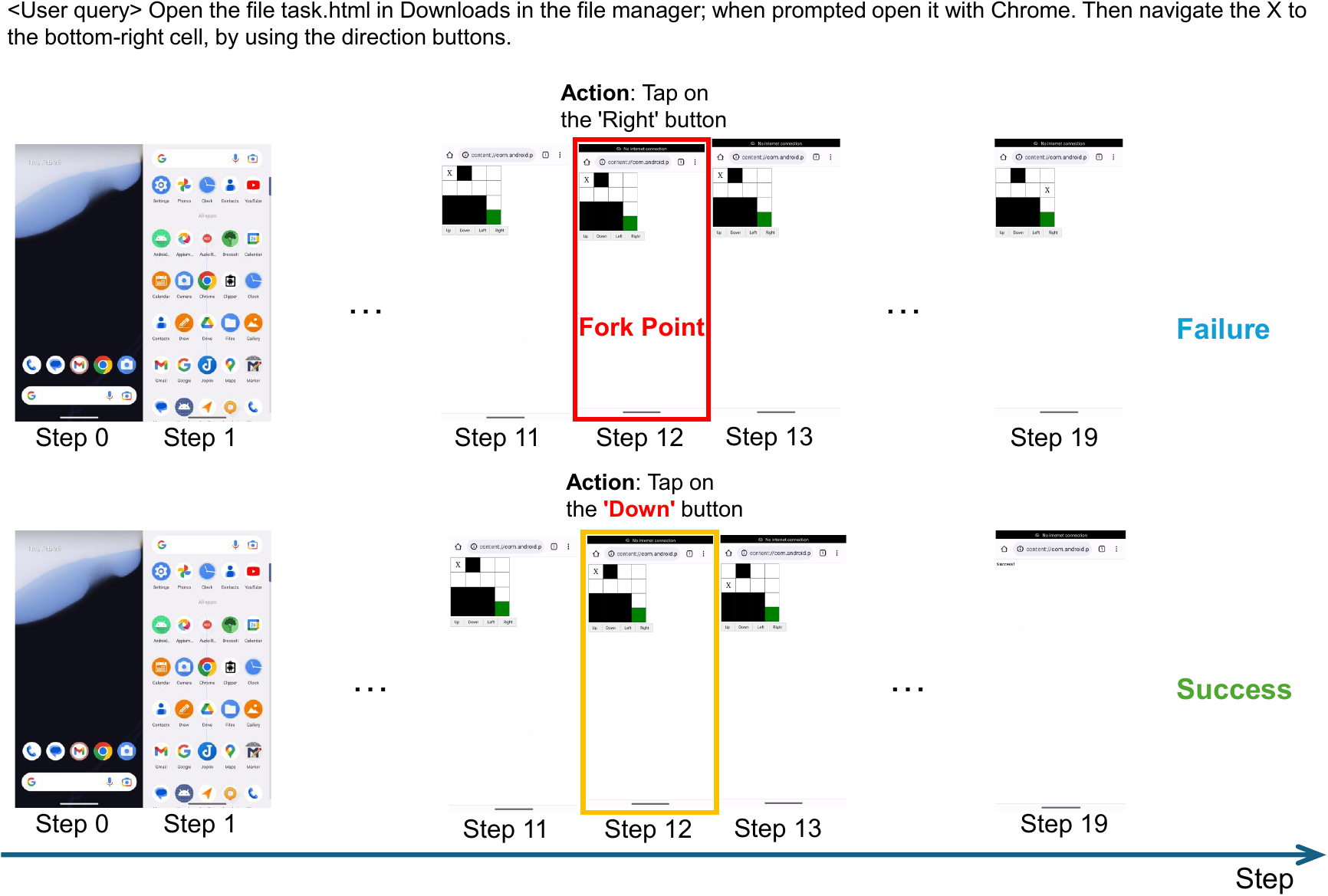}
    \caption{\textbf{Example of fork point detection on \emph{BrowserMaze} task.} Both failed and successful trajectories share the same screen state at Step 12 (fork point). The failed trajectory takes an invalid ``Right'' action (blocked by a wall), while the successful trajectory takes the correct ``Down'' action. The fork point detection mechanism identifies this divergence and uses the correct action from the successful trajectory to supervise the failed one at this critical step.}
    \label{fig:browsermaze}
\end{figure}

\paragraph{Rejection Fine-Tuning} RFT exhibits a robust and consistent capability to enhance agent performance through iterative optimization. As illustrated in Figure~\ref{fig:self_evolving} (Left), the Qwen3-VL-4B-Instruct model shows substantial gains in both Pass@1 and Pass@k metrics across four rounds of RFT. Notably, it demonstrates continued, steady improvements that highlight the efficacy of the self-evolving process. Figure~\ref{fig:self_evolving} (Right) shows the training curves for GRPO and PPO when initialized directly from the Qwen3-VL-4B-Instruct model. Both RL methods exhibit slow performance improvements, taking approximately 175 steps to reach the performance of a single RFT iteration (64.0\%). These results suggest that directly applying RLVR training from the base model is highly inefficient, highlighting the importance of using RFT as a reliable initialization strategy—providing a necessary ``warm-start'' for mastering complex, long-horizon agentic tasks. We adopt the model from the third iteration (Pass@1 = 73.2\%) as the foundation for subsequent agentic multi-turn RL training (GRPO and PPO) and our proposed GRSD. \looseness-1

\begin{figure}[t]
    \centering
    \includegraphics[width=0.97\linewidth]{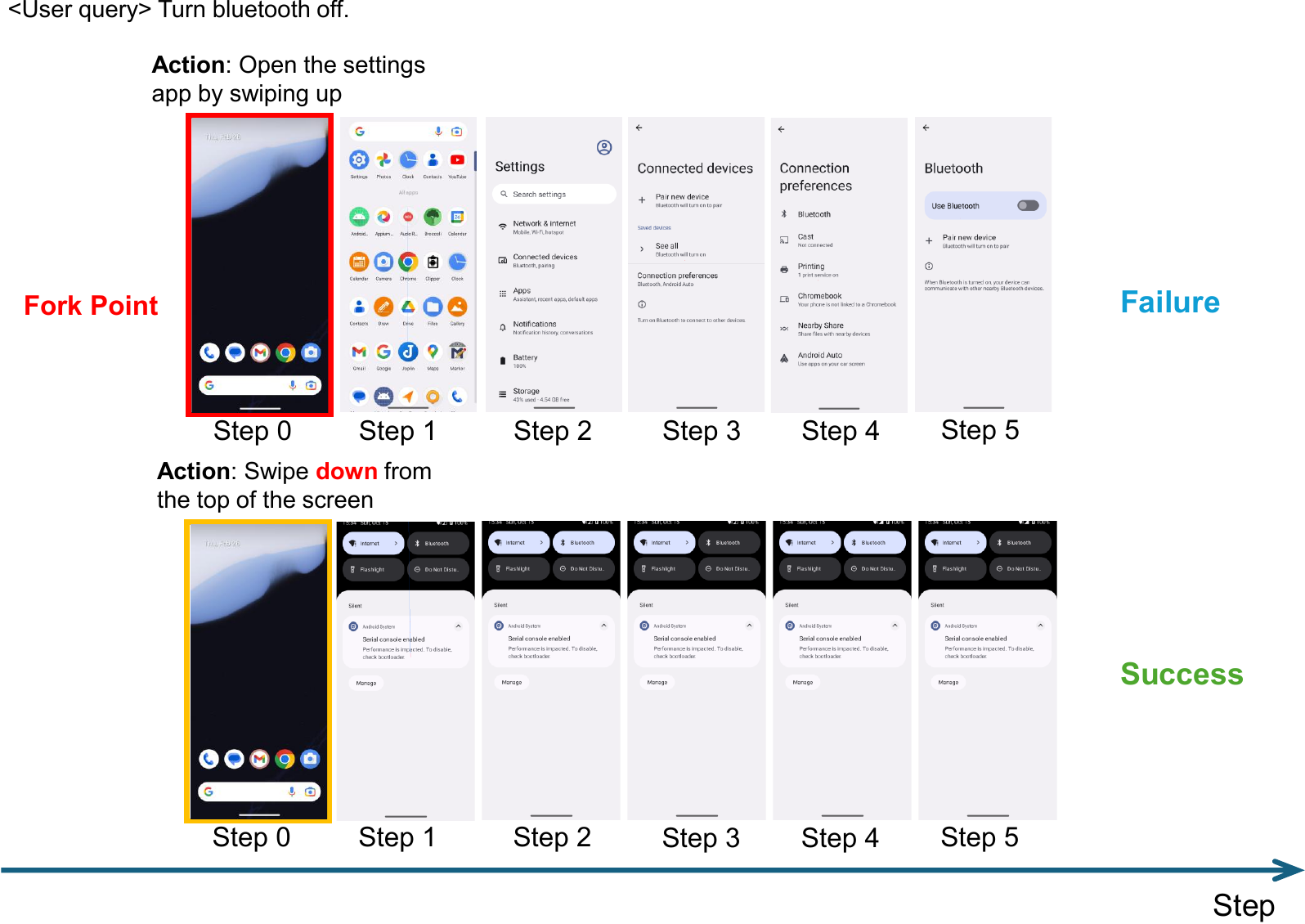}
    \caption{\textbf{Example of fork point detection on \emph{SystemBluetoothTurnOff} task.} The fork point occurs at Step 0 (initial state), where both trajectories start from the same home screen. The failed trajectory attempts to open the settings app with an incorrect upward swipe, while the successful trajectory uses a downward swipe to open the notification shade and access quick settings. Fork point detection identifies this initial divergence, providing corrective supervision at the very first step.}
    \label{fig:bluetooth}
\end{figure}

\paragraph{Fork Point Detection} To intuitively validate the functionality and practical value of the fork point detection mechanism, we demonstrate how it works by visualizing successful and failed trajectories of two representative tasks in AndroidWorld, \emph{BrowserMaze} and \emph{SystemBluetoothTurnOff}, as depicted in Figure \ref{fig:browsermaze} and \ref{fig:bluetooth}, respectively. These visualization results capture the critical divergence moments between failed and successful rollouts, directly illustrating how our method identifies shared screen states and corrects erroneous actions at key states.

To be specific, in the \emph{BrowserMaze} task (Figure \ref{fig:browsermaze}), both the failed and successful trajectories share an identical screen state at Step 12 (the fork point). The agent incorrectly taps the ``Right'' button in the failed trajectory, which is invalid (since there is a wall on the right side) and leads to task failure. The successful trajectory takes the correct ``Down'' action at Step 12 and ends up completing the task. By identifying this fork point (Step 12), our method extracts the correct action from the successful trajectory to supervise the failed one at this exact step. This provides efficient guidance to the agent on long-horizon navigation tasks like this, enabling the agent to avoid invalid moves and hence improving the task success rate.

Another interesting fact is that the fork point can be either a middle state or the initial state. As shown in the \emph{SystemBluetoothTurnOff} task (Figure \ref{fig:bluetooth}), the fork point occurs at Step 0, where both trajectories start from the same home screen. The failed trajectory intends to open the settings app via an incorrect upward swipe, while the successful trajectory uses a downward swipe to open the notification shade and access the quick settings. Fork point detection isolates this initial divergent action, providing valuable corrections at the starting stage. Note that this task requires the agent to turn on the bluetooth first and then turn it off (the default status of the bluetooth is off). These examples demonstrate that the fork point detection mechanism is both necessary for effectively learning from failed trajectories and for providing dense, step-level supervision for mobile GUI agent training.

\begin{figure}[t]
    \centering
    \includegraphics[width=0.97\linewidth]{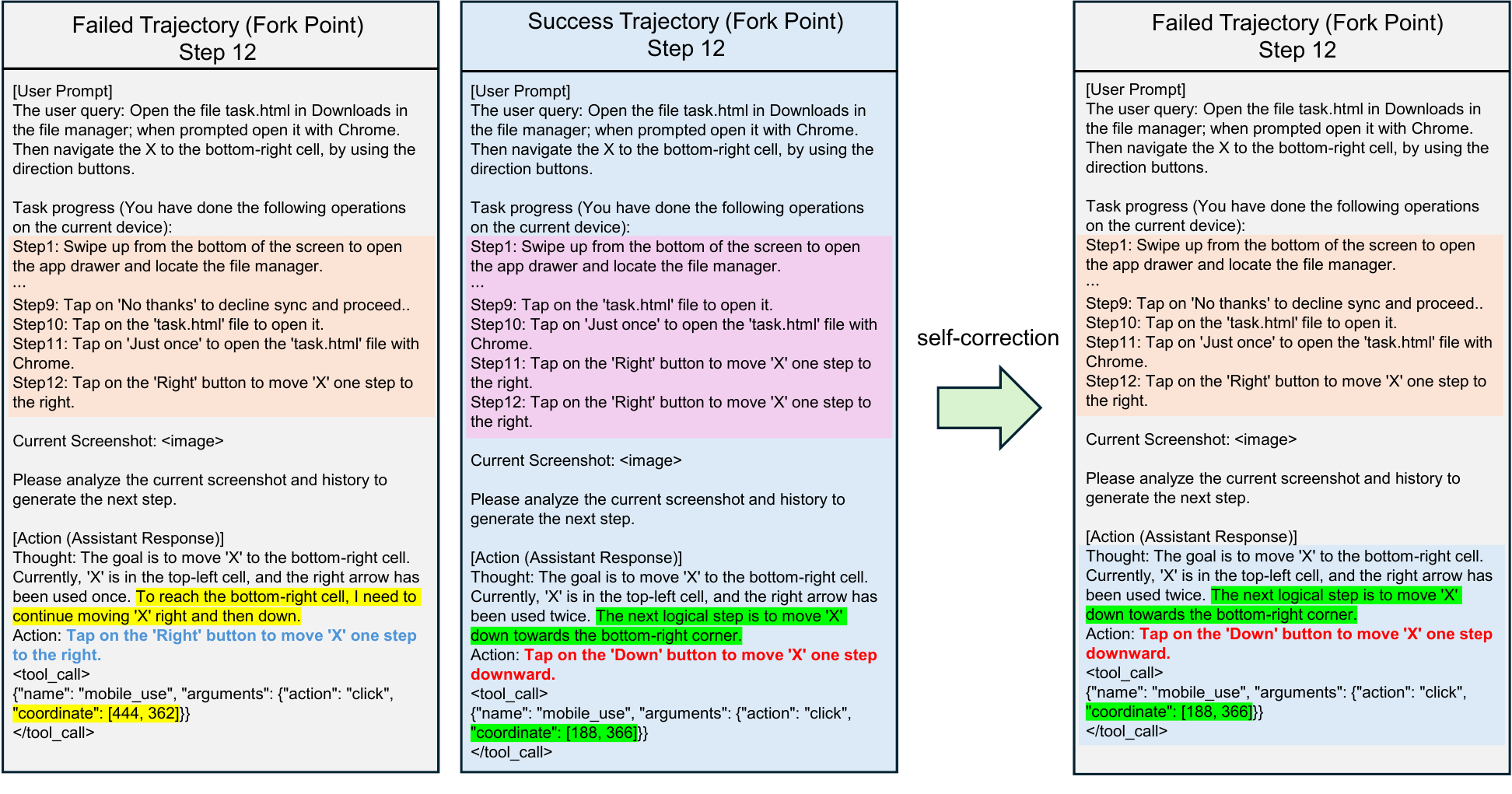}
    \caption{\textbf{Illustration of the self-corrective samples.} Both failed and successful trajectories share the same screen state at Step 12 (fork point). The failed trajectory incorrectly takes the ``Right'' action with flawed reasoning (``I need to continue moving `X' right and then down''), while the successful trajectory correctly takes the ``Down'' action with proper reasoning (``The next logical step is to move `X' down towards the bottom-right corner''). Fork point detection identifies this divergence, enabling the construction of self-corrective samples that transform failed trajectories into high-quality supervised data.}
    \label{fig:selfcorrectivesamples}
\end{figure}

\paragraph{Self-Corrective Sample} We now demonstrate how \ours performs self-correction at those fork points. We use the \emph{BrowserMaze} task as the example, with its visualized successful and failed rollouts shown in Figure \ref{fig:browsermaze}. The fork point occurs at Step 12 where both the successful and failed trajectories share the same screen state but diverge in their action selection. We show in Figure \ref{fig:selfcorrectivesamples} the construction process of the self-corrective samples. In the failed trajectory, the agent incorrectly reasons that \emph{``To reach the bottom-right cell, I need to continue moving `X' right and then down"} and taps the ``Right" button—an invalid move blocked by a wall. In contrast, the successful trajectory correctly reasons that \emph{``The next logical step is to move `X' down towards the bottom-right corner"} and takes the ``Down” action. By extracting the correct thinking process, the executed action, and the tool call process, we construct a self-corrective sample that transforms the failed trajectory into high-quality supervised data. This also enables the mobile GUI agent to effectively learn from the failed experiences and improve its decision-making capability without expensive human annotations.

\begin{figure}[t]
    \centering
    \includegraphics[width=0.49\linewidth]{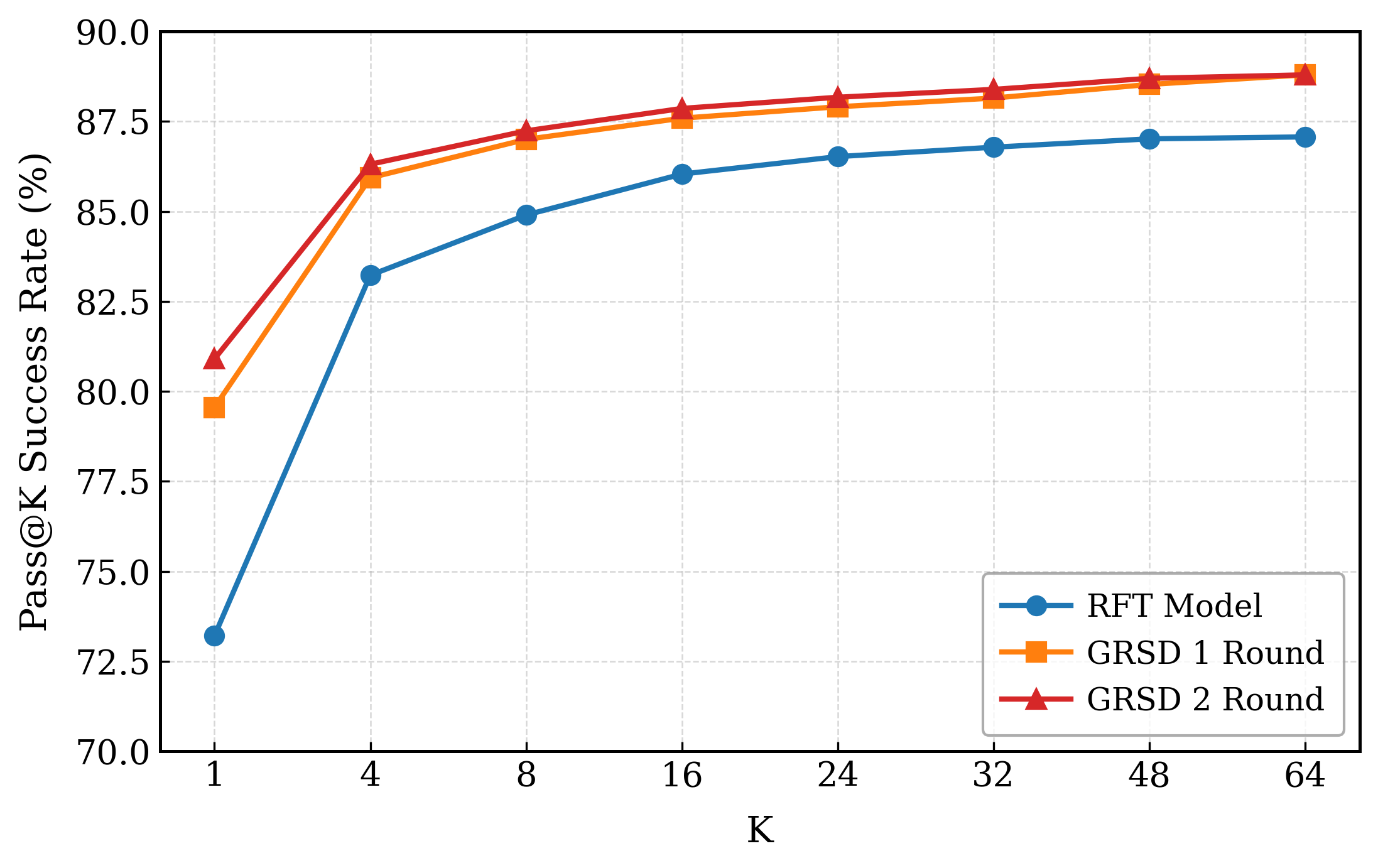}
    \includegraphics[width=0.49\linewidth]{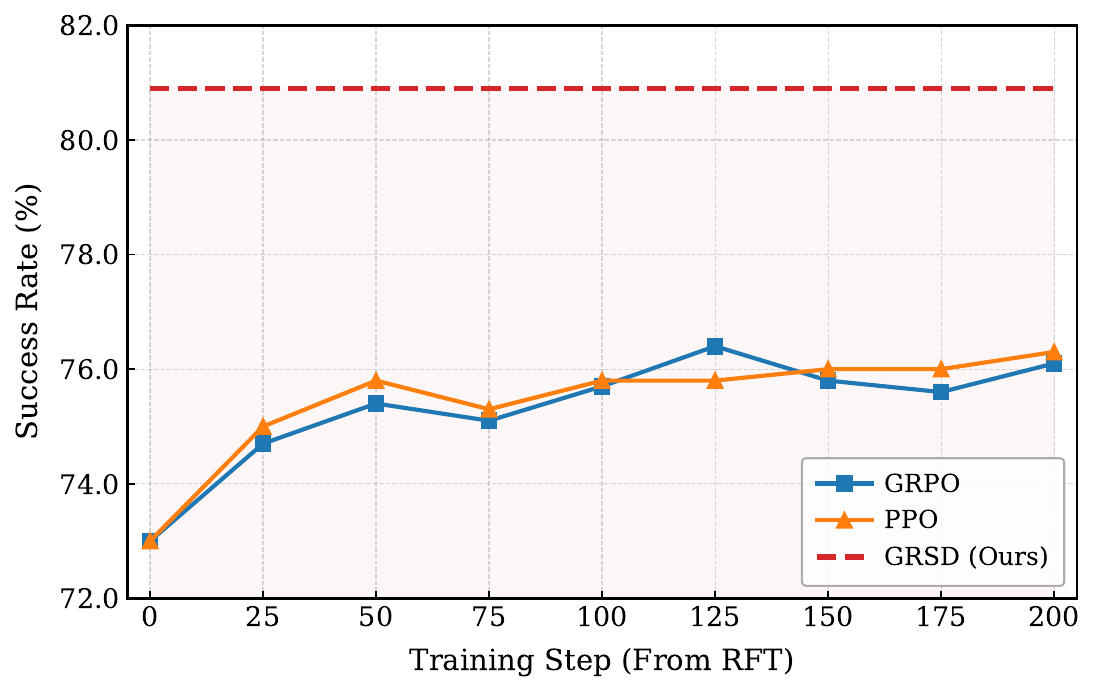}
    \caption{\textbf{Training performance comparison of GRSD, GRPO, and PPO.} All methods start from the same RFT model with 73.2\% success rate. GRSD successfully boosts the agent's Pass@1 performance to 81\% (Left), while GRPO and PPO show slower progress and plateau around 76\% (Right). The results demonstrate that GRSD's fork point detection and self-correction mechanisms enable more effective learning compared to standard RL baselines.}
    \label{fig:training}
\end{figure}

\paragraph{Comparison with RL Methods} We evaluate the performance of GRSD against GRPO and PPO. The results are presented in Fig.~\ref{fig:training}. We find that while all methods originate from the same RFT model with 73.2\% success rate, GRSD significantly boosts success rate to 81\%, whereas GRPO and PPO exhibit sluggish progress and eventually plateau at 76\%. This substantial performance gap stems from two primary limitations in standard RL baselines: first, the lack of an effective credit assignment mechanism in long-horizon GUI agent tasks obscures the identification of critical decision steps; second, these methods struggle to distill actionable insights from failed trajectories for rapid improvement. In contrast, GRSD leverages precise fork point detection to pinpoint decisive actions and employs self-distillation for rapid error correction, thereby facilitating a highly effective self-evolution process.

\begin{figure}[t]
    \centering
    \includegraphics[width=0.97\linewidth]{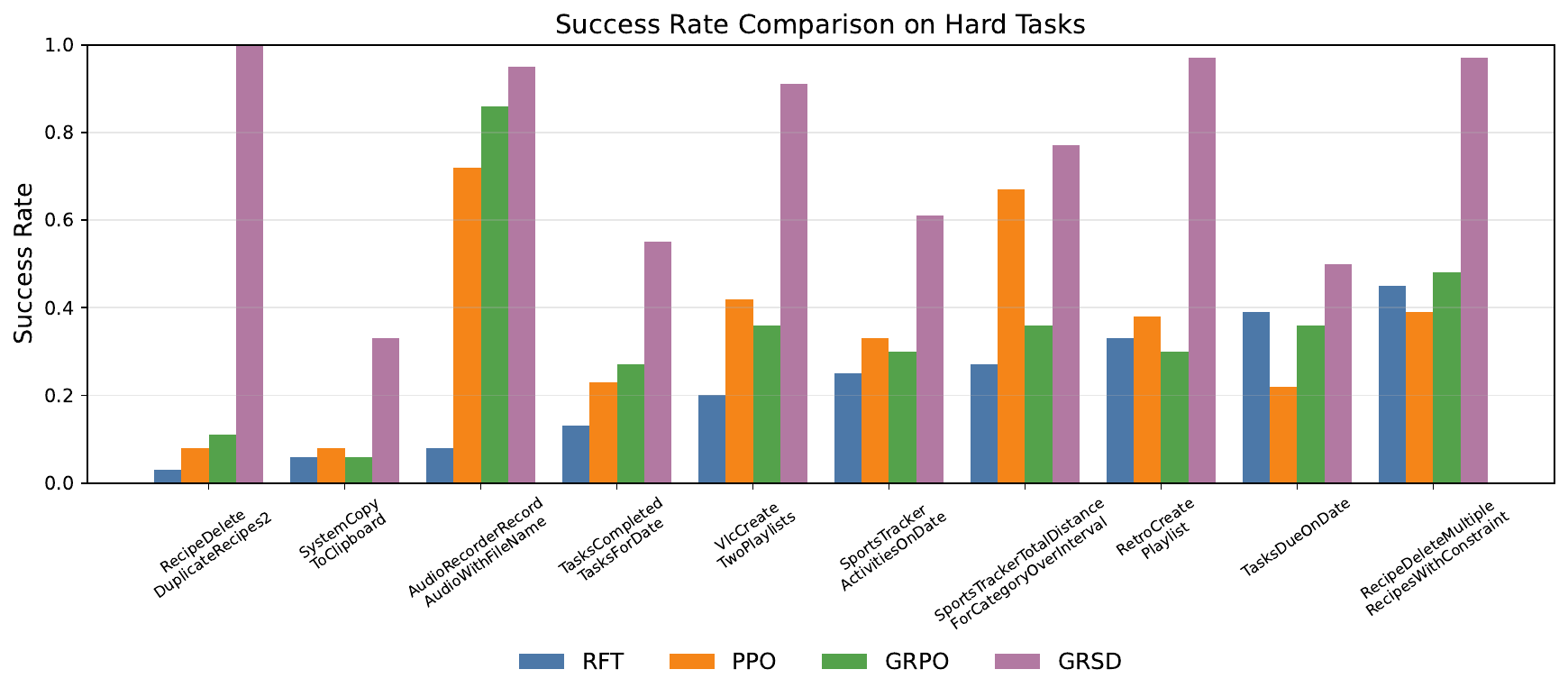}
    \caption{\textbf{Performance comparison of GRSD, GRPO, and PPO on ten representative low-success-rate tasks.} GRSD consistently achieves the highest success rate across all tasks, significantly outperforming both PPO and GRPO. In contrast, PPO and GRPO struggle to make substantial gains due to the scarcity of successful samples and the lack of effective credit assignment mechanisms. These results demonstrate that GRSD enables efficient learning from failed trajectories even in sparse-reward environments.}
    \label{fig:GRSD_on_low_success_rate}
\end{figure}

\paragraph{Effectiveness of GRSD} To evaluate whether GRSD can effectively learn from failure trajectories, we selected ten representative tasks where the baseline RFT model exhibited the lowest success rates. As illustrated in Fig.~\ref{fig:GRSD_on_low_success_rate}, both PPO and GRPO struggle to achieve significant performance gains on these challenging tasks. This stagnation is primarily due to the scarcity of successful samples during exploration, coupled with the absence of robust credit assignment and error correction mechanisms at critical decision points. In contrast, GRSD identifies pivotal decision junctures within failed trajectories and leverages successful counterparts through self-distillation. This approach substantially enhances the success rate across these low-performing tasks, demonstrating an efficient self-evolving capability even in sparse-reward environments.

\section{Discussion}
\paragraph{Real-time execution and SSIM-based matching.}
Following AndroidEnv~\cite{toyama2021androidenv}, mobile GUI interaction is inherently asynchronous: observations are streamed at a device-dependent frame rate, actions are executed with non-negligible latency, and the OS continues evolving while the agent is deciding. This real-time property can affect SSIM-based fork-point detection in two major ways. First, \emph{temporal misalignment}: two trajectories may correspond to the same logical UI state but be captured at slightly different moments (e.g., during animations, keyboard transitions, or loading), which can lower SSIM and lead to missed matches. Second, \emph{transient visual perturbations}: dynamic widgets such as cursor blinking, toast notifications, progress indicators, and clock updates can alter local pixels without changing the semantic state, introducing noisy high-SSIM or low-SSIM pairs. Therefore, under streaming observations, SSIM should be treated as a strong but imperfect proxy for state equivalence. \looseness-1

\paragraph{Practical implications for GRSD.}
A practical direction is to make matching explicitly time-aware rather than relying on single-frame comparison. Concretely, instead of matching one failed frame to one successful frame, we can match over a short temporal window and keep the best candidate, optionally with temporal smoothness constraints across neighboring steps. We can further reduce noise by masking high-variance UI regions (e.g., status bar, keyboard pop-ups, transient overlays) and combining SSIM with lightweight structure signals such as OCR/layout tokens or accessibility-tree cues. These additions keep the method efficient while improving robustness to asynchronous execution artifacts, which is particularly important when transferring from offline traces to real-time deployment settings.

\paragraph{Limited action space.}
Another practical limitation is the predefined action space used in AndroidWorld (Table~\ref{tab:action_space}). While high-level primitives (e.g., click, swipe, type, and navigation actions) make data generation and verification tractable, they abstract away low-level touch dynamics emphasized by AndroidEnv~\cite{toyama2021androidenv}, where raw interactions are continuous and asynchronously interpreted by the OS. This abstraction reduces exploration difficulty and improves training stability, but may also underestimate real-world failure modes related to gesture duration, trajectory shape, release timing, and actuation noise. Consequently, policies trained in a limited action space can be less robust when transferred to settings with finer-grained controls or different action wrappers. A promising direction is hierarchical action modeling: retain high-level actions for sample-efficient learning, then introduce low-level gestures and perturbations during post-training to improve transfer robustness.

\section{Conclusion}

In this paper, we address two critical challenges in mobile GUI agent training: inefficient learning from failed trajectories and ambiguous credit assignment under sparse rewards. As a result, we propose \ours trained by a two-stage self-evolving framework consisting of Rejection Fine-Tuning (RFT) and Group Relative Self-Distillation (GRSD). RFT enables automatic data–model co-evolution, while GRSD leverages fork point detection to provide dense step-level supervision from successful trajectories. Experimental results on 116 AndroidWorld tasks show that our 4B model achieves an 81.0\% pass@1 success rate across 116 AndroidWorld tasks, outperforming all baselines (with larger model sizes) and human-level performance. Ablation and case studies validate the effectiveness of our core designs. For future work, it would be interesting to extend the \ours framework to other GUI tasks (we only consider AndroidWorld in this work), explore more adaptive reasoning and self-correction mechanisms, and further improve the efficiency and robustness of mobile GUI agents in real-world scenarios.

\bibliographystyle{iclr2025_conference}
\bibliography{main}



\end{document}